\documentclass[10pt,twocolumn,letterpaper]{article}

\usepackage[pagenumbers]{cvpr}
\usepackage{times}
\usepackage{epsfig}
\usepackage{graphicx}
\usepackage{amsmath}
\usepackage{amssymb}

\usepackage{float}

\usepackage[table]{xcolor}
\usepackage{colortbl}
\usepackage{booktabs, makecell}
\usepackage{footmisc}
\usepackage[pagebackref=false,breaklinks=true,colorlinks,bookmarks=false]{hyperref}
\usepackage{cleveref}
 \usepackage[verbose]{placeins} 
 \usepackage{multirow}
 \usepackage{enumitem}

\usepackage[compress]{cite}

\usepackage{soul}

\usepackage{tabularx}
\usepackage{nicematrix}

\usepackage[accsupp]{axessibility}

\usepackage{makecell}

\usepackage{pifont}

\usepackage{arydshln}   % dashed horizontal/vertical lines

\usepackage{tikz}
\usetikzlibrary{patterns}
\usetikzlibrary{patterns.meta}
\definecolor{grpA}{RGB}{166,206,227}
\definecolor{grpB}{RGB}{178,223,138}

\usepackage{caption}

\newcommand{\dashmidrule}{%
  \noalign{\vskip\aboverulesep}%
  \hdashline[2pt/5pt]%
  \noalign{\vskip\belowrulesep}%
}

\usepackage{fancyhdr}
\fancypagestyle{firststyle}
{
   \fancyhf{}
   
   \fancyhead[C]{
    \textcolor{red}{
       The content of this paper was presented at IJCB 2025. If you find this research useful, please cite the following paper: \\ Ž. Babnik, D.K. Jain, P. Peer, V. Štruc, "FROQ: Observing Face Recognition Models for Efficient Quality Assessment" in Proceedings of the IEEE International Joint Conference on Biometrics (IJCB), 2025.}
    }
}

\begin{document}

%%%%%%%%% TITLE
\title{\vspace{-3mm}
    FROQ$^{\ref{fnote:froq}}$: Observing Face Recognition Models for Efficient Quality Assessment\vspace{-2mm}%Quality Assessment through Observation of Face Recognition 
%Face Recognition Models are Quality Estimators
}

\author{Žiga Babnik$^{1}$, Deepak Kumar Jain$^2$, Peter Peer$^1$, Vitomir Štruc$^1$\\
$^1$University of Ljubljana, Ljubljana, Slovenia\\
$^2$Dalian University of Technology, Dalian, China\\
%{\tt\small \{ziga.babnik$^1$,vitomir.struc$^4$\}@fe.uni-lj.si,dkj@ieee.org$^2$,peter.peer@fri.uni-lj.si$^3$}
%\and
%Deepak Kumar Jain\\
%Dalian University of Technology\\
%{\tt\small dkj@ieee.org}
%\and
%Peter Peer\\
%University of Ljubljana, Faculty of Computer and Information Science\\
%{\tt\small peter.peer@fri.uni-lj.si}
%\and
%Vitomir Štruc\\
%University of Ljubljana, Faculty of Electrical Engineering\\
%{\tt\small vitomir.struc@fe.uni-lj.si}
}

%\maketitle
%\thispagestyle{empty}

\iffalse
\twocolumn[{%
  \renewcommand\twocolumn[1][]{#1}%

  \vspace{5mm}
  \thispagestyle{firststyle}
  \maketitle
  \thispagestyle{firststyle}

  \begin{center}
    \vspace{-6mm}
    \includegraphics[width=.78\textwidth, trim=0 0 0 0, clip]{figures/teaser.pdf}\vspace{-4mm}

    % Legend row under the image (no amsmath \text{})
    {\small\flushleft\hspace{15mm}Input Samples\hspace{30mm}Face Recognition Model\hspace{34mm}Quality Estimates\par}\vspace{2mm}

    \captionof{figure}{\textbf{Illustration of the concept behind the proposed FROQ$^\ref{fnote:froq}$ technique.}
    Face Recognition (FR) models condense face samples into feature vectors. In the process, they encode identity-specific information, but also other non-identifying cues, such as face-sample quality~\cite{serfiq, faceqan, grafiqs}.
    Unsupervised Face Image Quality Assessment (FIQA) techniques can extract quality information directly from FR models, but incur a high computational cost. Supervised techniques are efficient, but typically require extensive training with complex loss functions and dedicated (FIQA) model architectures.
    FROQ combines the best from both worlds and efficiently estimates face-image quality using only the given FR model, by observing a set of specific, carefully chosen intermediate representations, while avoiding costly training and the reliance on custom FIQA-model architectures.}
    \label{fig:teaser}
  \end{center}%
}]
\fi

\twocolumn[{%
  \renewcommand\twocolumn[1][]{#1}%

  \vspace{5mm}
  \thispagestyle{firststyle}
  \maketitle
  \thispagestyle{firststyle}

  % --- Single minipage keeps grouping tidy and avoids fragile centers ---
  \begin{minipage}{\textwidth}
    \centering
    \vspace{-6mm}

    \includegraphics[width=.78\textwidth, trim=0 0 0 0, clip]{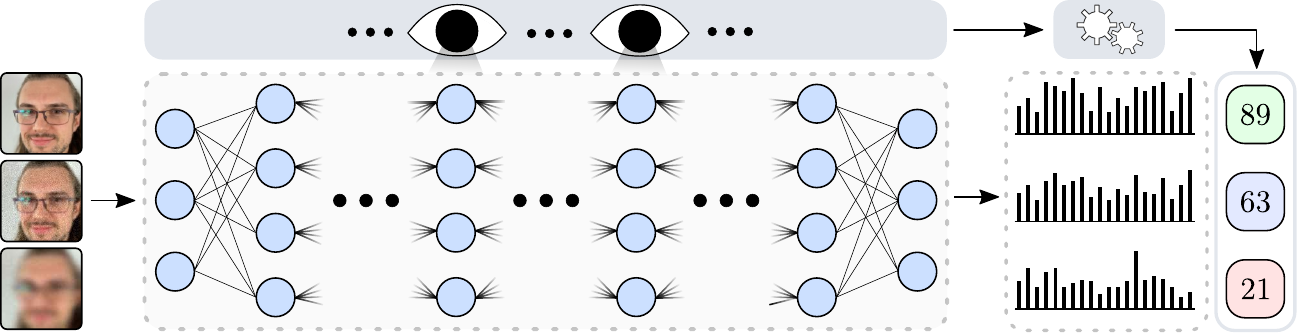}\vspace{-4mm}

    % One-line legend (no amsmath \text, no extra environments)
    {\small
    \vspace{3mm}
      \hspace{15mm}Input Samples\hfill
      Face Recognition Model\hfill
      Quality Estimates\hspace{22mm}
    }\vspace{2mm}

    % Caption + label
    \captionof{figure}{\textbf{Illustration of the concept behind the proposed FROQ$^{\ref{fnote:froq}}$ technique.}
    Face Recognition (FR) models condense face samples into feature vectors, encoding identity-specific information and non-identifying cues such as face-sample quality~\cite{serfiq, faceqan, grafiqs}.
    Unsupervised FIQA methods can extract quality directly from FR models but are computationally heavy; supervised methods are efficient but typically require complex training and custom architectures.
    FROQ efficiently estimates face-image quality using only the given FR model by observing carefully chosen intermediate representations, avoiding costly training and dedicated FIQA architectures.
    }
    \label{fig:teaser}
  \end{minipage}%
}]

%%%%%%%%% ABSTRACT
\begin{abstract}
\vspace{-2mm}
    Face Recognition (FR) plays a crucial role in many critical (high-stakes) applications, where errors in the recognition process can lead to serious consequences.~Face Image Quality Assessment (FIQA) techniques enhance FR systems by providing quality estimates of face samples, enabling the systems to discard samples that are unsuitable for reliable recognition or lead to low-confidence recognition decisions. Most state-of-the-art FIQA techniques rely on extensive supervised training to achieve accurate quality estimation. In contrast, unsupervised techniques eliminate the need for additional training but tend to be slower and typically exhibit lower performance. In this paper, we introduce FROQ\footnote{\label{fnote:froq} \textbf{FROQ} is pronounced as \textit{FROG.}} (\textbf{F}ace \textbf{R}ecognition \textbf{O}bserver of \textbf{Q}uality), a semi-supervised, training-free approach that leverages specific intermediate representations within a given FR model to estimate face-image quality, and combines the efficiency of supervised FIQA models with the training-free approach of unsupervised methods.~A simple calibration step based on pseudo-quality labels allows FROQ to uncover specific representations, useful for quality assessment, in any modern FR model.~To generate these pseudo-labels, we propose a novel unsupervised FIQA technique based on sample perturbations.
    %Most state-of-the-art FIQA techniques require extensive supervised training to estimate a sample's quality accurately. Unsupervised techniques do not require additional training, but are, as a result, slower and do not perform as well. In this paper, we present FROQ\footnote{\label{fnote:froq} \textbf{FROQ} is pronounced as \textit{FROG.}}, a semi-supervised, training-free method, based on observations of specific intermediate representations within a given FR model. A simple calibration step, using pseudo-quality labels, allows FROQ to uncover specific representations, useful for quality assessment, in any modern FR model. To obtain the labels, we present a novel unsupervised FIQA technique, based on sample perturbations.   % leverages a small set of quality-labeled images to infer which intermediate representations, within the FR model, are useful for quality assessment. The quality score is obtained by simply aggregating and averaging the values of all observed intermediate representations.  %Once the informative representations have been discovered, quality scores can be computed by applying an aggregation function to the values of the specified intermediate values. 
    %The method can quickly and accurately asses the quality of a sample, without the need for any supervised training or additional forward or backward passes.
    Comprehensive experiments with four state-of-the-art FR models and eight benchmark datasets show that FROQ leads to highly competitive results compared to the state-of-the-art, achieving both strong performance and efficient runtime, without requiring explicit training. %~The code for FROQ will be made publicly available after review.\vspace{-4mm}
    The code for FROQ is available from: {\small \url{https://github.com/LSIbabnikz/FROQ}}
    \vspace{-3mm}

    % (MADE BY CHATGPT NOT FOR USE IN FINAL) Face Recognition (FR) systems critically rely on accurate assessment of face image quality to maintain robust performance in real-world scenarios. This paper introduces FROQ (Face Recognition Observer of Quality), a novel face image quality assessment (FIQA) method that efficiently estimates image quality through observations of intermediate representations during the face recognition process. Unlike existing supervised methods that require extensive training or unsupervised methods that incur significant computational overhead, FROQ uniquely leverages a semi-supervised strategy by selecting informative intermediate representations using a small calibration dataset. By employing a single forward pass through the FR model and a carefully designed aggregation function, FROQ achieves high accuracy in quality estimation without additional supervised training or parameter adjustments. Extensive experiments demonstrate that FROQ outperforms state-of-the-art unsupervised techniques and closely matches the best supervised approaches across multiple datasets and FR models, achieving superior runtime efficiency and robustness.

\end{abstract}

%%%%%%%%% BODY TEXT
\section{Introduction}\label{sec:introduction}\vspace{-1mm}

Face Recognition~(FR) is an important research area with numerous real-world applications in security and surveillance, border control, police investigations, online banking, and mobile applications, among others \cite{fr_survey}. The reliability of FR models in these applications is critical, as errors in the recognition process can compromise user privacy, result in monetary loss, or even lead to legal consequences. While significant advances have been made in FR technology over the years, FR systems still fail to accurately determine identity when deployed in challenging  acquisition conditions \cite{grm2018strengths,fr2,fr_synth}, where variations in pose, illumination, or other environmental factors cannot be controlled for. To mitigate these issues, FR models often incorporate Face Image Quality Assessment (FIQA) techniques with the goal of assessing the fitness of the input images for recognition.

%either due to unconstrained variations in the face pose and illumination, or because of other environmental variables not accounted for in the capture process. To combat the issues occurring during the capture of samples, FR models employ Face Image Quality Assessment (FIQA) techniques.

%Face Recognition plays a key role in many important tasks, such as security systems, surveillance systems, online banking, etc. Such tasks rely on the performance and reliability of the underlying face recognition system, not making any mistakes regarding the end task of identification or verification, as each error could lead to hefty monetary loss and a potential breach of security. While some systems require cooperation from the subject when capturing an image, guaranteeing some lower bound on the quality of taken samples, not all systems do. In such cases, using Face Image Quality Assessment techniques proves to be an effective countermeasure to dealing with errors arising from low-quality captured samples. 

In accordance with ISO/IEC 29794-1~\cite{eval2}, modern FIQA techniques most often generate a \textit{unified quality score} that corresponds to the utility of the given face sample for the task of recognition. Here, the utility is typically measured by how likely the sample is to cause false-match errors during the recognition process. In this manner, samples less likely to cause false-match errors are considered to be of higher quality. The quality (or utility) estimates allow FR systems to reject or recapture samples below a certain quality threshold, improving the system's reliability. 

Existing FIQA techniques can be broadly categorized into: \textit{unsupervised} and \textit{supervised} methods. Unsupervised methods typically estimate sample quality by looking at the behavior of the FR model to %input~\cite{faceqan, diffiqa} (or intermediate) 
perturbations applied to the input face sample~\cite{faceqan, diffiqa, serfiq, iwbf2023}. Supervised methods, on the other hand, commonly train a quality-regression model, use pseudo-quality labels~\cite{pcnet, sdd-fiqa, lightqnet}, rely on a specific loss function~\cite{magface, adaface, ediffiqa}, or external (often generative) proxy tasks~\cite{faceqgen, diffiqa, clib-fiqa}.
Unlike supervised methods, unsupervised techniques are easily adapted to any target FR model, but they are noticeably slower when assessing quality, as they require several forward or even additional backward passes through the target FR model. Supervised methods are more efficient during inference, but often rely on dedicated model architectures and, hence, require more work to be adapted for a specific target FR model.% On the other hand, supervised methods require training an auxiliary quality regression model or fine-tuning existing FR models. They can also require the use of specific loss functions or pseudo-quality labels. This, however, allows them to compute the quality score using only a single pass through the FR or auxiliary quality estimation model.

In this paper, we present a novel quality assessment technique, called \textbf{FROQ} (\textbf{F}ace \textbf{R}ecognition \textbf{O}bserver of \textbf{Q}uality), capable of accurately estimating face-sample quality that needs only a single forward pass through the FR model, as shown in Figure~\ref{fig:teaser}. The method can be easily adapted to any FR model and requires no supervised training or additional parameters to tune. The \textbf{main contribution} of the approach \textbf{is the Quality Observer}, whose goal is to closely monitor specific intermediate representations produced by the FR model during the recognition process. These representations are used as is to estimate the final quality score for a given input face sample. To discover useful intermediate representations, we present a simple semi- (or weakly) supervised approach, which evaluates the usefulness of individual representations for the task of quality estimation through the use of a small quality-labeled calibration set. In this way, FROQ combines the characteristics of both supervised and unsupervised techniques, achieving excellent runtime, estimating the quality within a single forward pass, without the need for any supervised training or additional FR-model parameters.

%Our contributions:
%\begin{enumerate}
%    \item 
%    \item
%\end{enumerate}

\begin{figure*}[!ht]
    \centering
    \includegraphics[width=0.93\linewidth]{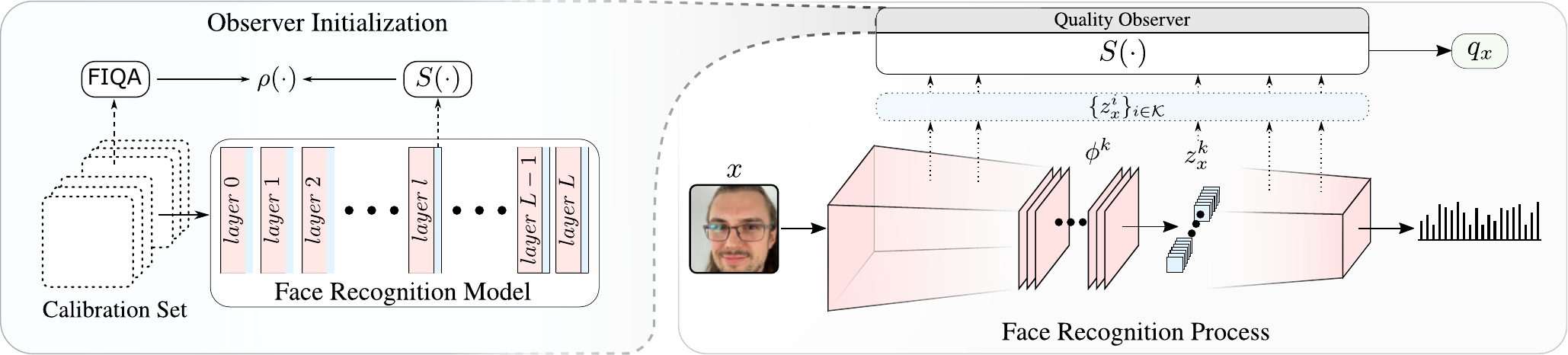}
    \caption{\textbf{High-level overview of FROQ.} FROQ estimates sample quality by observing specific intermediate representations produced by the recognition process. The quality score $q_x$ for a given sample $x$ is computed by applying an aggregation function $S(\cdot)$ on the values of the observed intermediate representations $\{z^i_x\}_{i\in\mathcal{K}}$. The set of intermediate representations $\mathcal{K}$ observed during the recognition process is determined in the \textit{Observer Initialization} step. Here, each layer of a given FR model is evaluated using a calibration set of images and their corresponding quality scores obtained through an auxiliary FIQA approach, using Spearman's rank correlation $\rho(\cdot)$.\vspace{-2mm}
    }
    \label{fig:method_overview}
\end{figure*}

\section{Related Work}\label{sec:related_work}

In this section, we provide a brief overview of relevant work on face image quality assessment and discuss both \textit{unsupervised} and \textit{supervised} FIQA techniques. For a more comprehensive coverage of the topic, please see~\cite{survey}. 

\vspace{0.9mm}\noindent\textbf{Unsupervised Methods.}\label{sec:related_work:subsection:unsupervised_methods}
Unsupervised FIQA methods do not require any supervision when building FIQA models. Instead, they commonly estimate sample quality by observing the effects of various perturbations on the sample's representation within the latent space of the target FR model. One of the earliest methods, SER-FIQ~\cite{serfiq}, applied dropout to intermediate representations to estimate sample quality. The dropout layer removes certain values from the representation and can be seen as a type of random occlusion on the latent representation. FaceQAN~\cite{faceqan} proposed an adversarial attack to predict quality. More specifically, it used the noise applied to the sample during an (adversarial) attack as the perturbation of choice. Recently, DifFIQA~\cite{diffiqa} introduced a combined approach, using two separate perturbations, encapsulated in the process of modern denoising diffusion probabilistic models (DDPMs). GraFIQs~\cite{grafiqs} presented a new type of unsupervised FIQA approach, focused on the statistics of the batch normalization layers during the backward pass through the target FR model. A common characteristic of unsupervised FIQA techniques is that they can typically be applied to any modern FR model without additional adjustments. However, they commonly require several forward or even backward passes to estimate quality, making them less computationally efficient. 

\vspace{0.9mm}\noindent\textbf{Supervised Methods.}\label{sec:related_work:subsection:supervised_methods}~Supervised FIQA techniques commonly require training (auxiliary) quality-regression networks or fine-tuning existing face recognition models to estimate face-sample quality and can be further subdivided by whether they need pseudo-quality labels or not.

Methods requiring pseudo-quality labels employ different annotation techniques to obtain the labeled data. The annotated data is then used to train quality regression models.~One of the earliest methods in this category, by Best-Rowden and Jain~\cite{bestrowden}, utilizes labels provided by human annotators, while FaceQNet, proposed by Hernandez-Ortega \textit{et al.}~\cite{faceqnet1, faceqnet2}, was among the first employing automatically generated pseudo-quality labels, computed by comparing input face samples to the highest-quality images (references) of the same identity.~A more robust approach, PCNet~\cite{pcnet}, used comparisons between several genuine (positive) samples to determine the pseudo labels. SDD-FIQA~\cite{sdd-fiqa} extended the idea presented by PCNet by also considering information from imposter image pairs. A similar approach was later used in LightQnet~\cite{lightqnet}, with an additional focus on minimizing the parameter count of the final quality extraction model. A quality-label optimization approach, applicable to any set of pseudo-quality labels, was proposed in eDifFIQA~\cite{ediffiqa}, leading to highly competitive results. Finally, CLIB-FIQA~\cite{clib-fiqa} presented a unique supervised technique that used labels of individual quality factors, such as blur, occlusion, lighting, \etc, in combination with the CLIP encoder~\cite{clip} to train a quality estimation model. While techniques from this group usually achieve good results, they need additional training on pseudo-quality labels and are limited by the expressiveness of the label generation process.

On the other hand, supervised FIQA methods that do not require pseudo-quality labels often use a custom loss function to train a model that can estimate both the feature and quality of a given face sample. For this reason, such methods are also called quality-aware FR methods. One of the earliest such methods, PFE~\cite{pfe}, learns a mean and variance vector, corresponding to the feature and quality of the given sample. MagFace~\cite{magface} proposes an extension of the popular ArcFace~\cite{arcface} loss function, with a magnitude-aware term, enabling the model to encode quality information into the magnitude of the feature vector. A special type of such a method, CR-FIQA~\cite{cr-fiqa}, uses information encoded in the feature space of a trained model to estimate the quality of samples. Specifically, it uses the Certainty-Ratio, which compares the distances of the given sample to the positive class center and the nearest negative class center.

\vspace{1mm}\noindent\textbf{Our Contribution.}\label{sec:related_work:subsection:supervised_methods}~The proposed FROQ technique leverages pseudo-quality labels; however, unlike supervised methods, the labels are not used to train a quality regressor. Instead, FROQ uses the labels to uncover intermediate representations of the target FR model that are useful for quality assessment. In this process, no new information is encoded into the FR model. For this reason, we categorize FROQ as a semi-supervised technique, combining characteristics of both unsupervised and supervised methods. It can compute the quality scores quickly, using a single forward pass through the FR model, without needing explicit training or limiting the target FR model to a specific loss.

%The proposed FROQ technique can be seen as semi-supervised, requiring a small set of quality-labeled face images to discover informative intermediate representations of the given FR model. No training or additional parameters are needed in the discovery process. Once the intermediate representations are uncovered, the approach can quickly (single forward pass) estimate the sample's quality. The method combines the efficient computation of quality scores, as seen in supervised methods, without needing supervised training or additional quality regression modules.

%The proposed FROQ method can be viewed as a mix between  tysupervised method, as it requires pseudo-quality labels. However, unlike previous supervised methods, which require quality labels, the focus of our method is not on the process of pseudo-quality label extraction. Instead, the main focus of our method is the use of pseudo-quality labels to leverage preexisting information already encoded in any modern FR model, for the task of face image quality assessment. Unlike all previous methods, the proposed technique does not need any additional training of FR models.

\section{Methodology}\label{sec:methodology}

The main goal of FR models is to extract identity information from any given face sample, in the form of a dense identity-information-rich representation called a feature vector. It has been shown that FR models also encode non-identifying information about the samples, such as face pose, expression~\cite{faceexpression}, as well as the quality of the sample~\cite{serfiq, sdd-fiqa, faceqan}. %To extract quality information, existing FIQA works require several forward or backward passes in the case of unsupervised methods~\cite{faceqan, serfiq, grafiqs, diffiqa},  supervised training or additional parameters, in the case of supervised methods~\cite{sdd-fiqa, cr-fiqa, magface}. 
In this section, we now introduce FROQ, a new semi-supervised FIQA approach that requires no training and can accurately predict sample quality using only a single forward pass through the target FR model. To estimate quality, FROQ uses a \textit{Quality Observer}, which tracks values of predetermined intermediate representations of the given FR model, as shown in Fig.~\ref{fig:method_overview}. The sample quality is then computed directly from the observed representations using a simple aggregation function. Both the set of observed intermediate representations and the aggregation function are determined in the \textit{Observer Initialization} step. %\vspace{-1mm}%To determine the important intermediate representations for observation, we introduce the \textit{Observer Initialization} step.

%Given previous work in the field of quality assessment of face images, it is clear that face recognition models or the process of face recognition somehow encodes information about the quality of the sample itself~\cite{faceqan, serfiq, grafiqs, cr-fiqa}. Methods use different strategies to extract this information from the model or face recognition process itself, by perturbing input samples or intermediate representations~\cite{faceqan, serfiq}, using statistics of the training process~\cite{cr-fiqa, sdd-fiqa}, or modifying the training loss of the model~\cite{magface, adaface}. All existing methods, to the best of our knowledge, employ additional training, or need several forward passes or even backward passes through the network, to estimate the quality of a given sample. Our method [Insert Name], however, estimates the quality by observing the interaction between the face recognition model and the input sample during a single forward pass. In other words the process of face recognition, provides our method with enough information to accurately estimate the quality of a face sample. The method relies on information stored in the pretrained face recognition model, more specifically in the individual learned layers and the intermediate representations produced by these layers. By observing the values of intermediate representations during recognition (a single forward pass) of a face sample, the method computes the final sample quality by aggregating a simple statistic computed over chosen intermediate layers. 

\subsection{Overview}\label{sec:methodology:subsection:overview}%\vspace{-0.2mm}

Assume a FR model $M$, which consists of $L$ layers and is parameterized by $\{\phi^l\}_{l=1}^L$, where $\phi^l$ are the parameters of the $l$-th layer. The goal of FROQ is to extract a quality estimate $q_x$ of the input face sample $x$ using the predefined \textit{Quality Observer}. Here, the observer determines the quality $q_x$ using a single forward pass of the sample $x$ through the recognition network $M$. It is defined by two dependent components: $(i)$ the aggregation function $S(\cdot)$, and $(ii)$ the set of intermediate representations $\mathcal{K}$. The role of the aggregation function is to map any intermediate representation $z^j$ into a single numerical value, where $z^j \in \{z^l\}_{l=1}^L$ is the intermediate representation produced by the $j$-th layer of the model $M$. The set $\mathcal{K}$ determines which representations will be observed and subsequently used for quality estimation. %\vspace{-0mm}%To determine both the aggregation function $S(\cdot)$ and the set $\mathcal{K}$, we perform the \textit{Observer Initialization} step.

\subsection{Observer Initialization}\label{sec:methodology:subsection:observer_definition}%\vspace{-1mm}

The quality observer is at the core of the proposed FROQ technique. During inference, it accurately determines the quality of any input sample by simply looking at specific intermediate layers. The values of the intermediate representations are first condensed into a single numerical score using a dedicated aggregation function, and then combined across different representations to compute the final quality estimate. In the following section, we describe how the two main components of the observer are determined, i.e., the aggregation function $S(\cdot)$ and the set of representations $\mathcal{K}$. % and applying an aggregation function to the values of the intermediate layers. To initialize it, the two main components, i.e., the aggregation function and the chosen intermediate representations, need to be defined. In the following section, we describe in detail the steps necessary to determine both components.

%The Quality Observer is at the core of our technique. During the face recognition process, it is tasked with assessing the quality of the processed face sample. To assess the quality, it looks at specific intermediate representations generated when running a face sample through the FR model, and joining them using an auxiliary function to form the final quality estimate. The set of intermediate representations monitored by the observer depends on the specified FR model and on the aggregation function $S(\cdot)$. The function $S(\cdot)$ is a simple function that maps an intermediate representation $Z_j$ into a single numerical value $o_j$. Since there are many possible such functions, the combined search space of possible aggregation functions and sets of intermediate representations is vast. For this reason, we decide to focus on discovering the best possible set of intermediate representations for a specific hand-crafted function $S(\cdot)$. 

\vspace{1.5mm}\noindent\textbf{The Aggregation Function.}\label{sec:methodology:subsection:observer_definition:subsubsection:aggregation_function}
The main goal of the aggregation function $S(\cdot)$ is to map any intermediate representation $z^l_x$ of the sample $x$ into a single numerical score. There exist infinitely many such functions that make evaluating all possible solutions impossible. Additionally, the choice of the aggregation function also affects the set of observed representations, which makes the combined search space of possible aggregation functions and sets of representations far too large to fully explore. For this reason, we hand-craft the aggregation function $S(\cdot)$ based on several insights about FR models, presented by prior works~\cite{magface, adaface, qaface, norm_extra}.

%Since there are many possible such functions, the combined search space of possible aggregation functions and sets of intermediate representations is vast. For this reason, we decide to focus on discovering the best possible set of intermediate representations for a specific hand-crafted function $S(\cdot)$.

Modern FR models are trained on medium-to-high quality samples~\cite{arcface, adaface, transface}, and so the learned parameters (filters) respond well to inputs corresponding to images of higher quality. This means that, generally, the amplitude of the individual layer's outputs (intermediate representations) should be correlated with the quality of the input samples. Following this insight, we design the aggregation function around the norm of intermediate values, formally: %. Formally, we define the aggregation function using:
\begin{equation}\label{equation:aggregation_function}
    S({z^l_x}) = \| \mathit{f}_{flatten}({z^l_x}) \|_{2},
\end{equation}
where $z^l_x$ is the intermediate representation produced by the $l$-th layer of the input sample $x$, $\mathit{f}_{flatten}(\cdot)$ is a flattening operation and $\| \cdot \|_{2}$ is the $\mathit{L}_2$ norm. Since the intermediate representation $z^l_x$ can be of arbitrary shape $(d_1, d_2, \dots, d_D)$, we first reshape it into a single dimension with  $(d_1 \times d_2 \times \dots \times d_D)$ elements, using $\mathit{f}_{flatten}(\cdot)$.

%The first defining feature of the observer is the aggregation function $S$, which takes as input any intermediate representation $z_i$ of the model $M$. A given intermediate representation $z_i^{d_1, d_2, \dots, d_n}$ can be defined in $n$-dimensions, where $n$ is typically $2$ or $3$, with arbitrary sizes of individual dimensions $d_k$, the function $S$ needs to be crafted with this in mind. Face Recognition models are trained on large datasets of mostly high quality face samples, therefore the majority of layers and individual filers are trained to maximally respond to high quality samples (and their intermediate representations), for this reason we base our function $S$ on the $L2$-norm of intermediate representations. The function $S$, used by the proposed method is defined as
%\begin{equation}\label{equation:aggregation_function}
%    S(z_i^{d_1, d_2, \dots, d_n}) = \| z_i^{d_1 \times d_2 \times \dots \times d_n} %\|_{2},
%\end{equation}
%where $\|\cdot\|_{2}$ denotes the $L2$-norm operation and $z_i^{d_1 \times d_2 \times \dots \times d_n}$ is the flattened intermediate representation $z_i$. 

\vspace{1.5mm}\noindent\textbf{Set of Intermediate Representations $\mathcal{K}$.}\label{sec:methodology:subsection:observer_definition:subsubsection:set_of_layers}
To determine the set of layers (representations) $\mathcal{K}$, we propose a simple semi-supervised approach, which evaluates the usefulness of individual layers for quality estimation.

%The second defining feature as previously stated is the set $\mathcal{K}$, which contains the indices of the chosen intermediate representations (and model layers) which best describe the quality of the input sample $x$, given a specific aggregation function $S$. To determine the set $\mathcal{K}$ we devise a two step process, in the first step each individual intermediate representation $z_k \forall k \in [0, K]$ is analyzed, then in the second step the set $\mathcal{K}$ is obtained by computing the best combinations of the top $l$ intermediate representations of the previous step. 

Given a calibration set of face images $\{x_i\}_{i=1}^{N}$, we compute pseudo-quality labels $\{\dot{q}_i\}_{i=1}^{N}$, using an auxiliary FIQA approach.~Here, any preexisting FIQA technique could be used. However, to separate the proposed method from previous works, we devise a custom FIQA approach based on face-sample perturbations, presented in Sec.~\ref{sec:methodology:subsection:custom_fiqa}. For a given FR model $M$, we investigate the quality information of each layer $l\in[1,L]$ within $M$ using:
\begin{equation}
    c^l = \rho(\{\dot{q}_i\}_{i=1}^{N}, \{\hat{q}^l_i\}_{i=1}^{N}),
\end{equation}
where $\rho(\cdot)$ is Spearman's rank correlation, and $\{\hat{q}^l_i\}_{i=1}^{N}$ the quality estimates obtained by applying the aggregation function $S(\cdot)$ on the intermediate representations produced by the $l$-th layer. The computed coefficient $c^l$ shows the similarity of the ranking of sample qualities provided by the auxiliary FIQA approach and the intermediate representation $l$. A higher correlation coefficient (close to $1$) shows that the two analyzed quality vectors have very similar quality rankings, indicating that the analyzed intermediate representations are suitable for the task of quality assessment. By evaluating all layers $l \in [1, L]$ in this manner, we obtain a set of layer correlation coefficients $\{c^l\}_{l=1}^{L}$. 

To determine the final set of intermediate layers $\mathcal{K}$, we take into account only the best $b$ layers, according to the correlation coefficients $\{c^l\}_{l=1}^{L}$ and collect these layers in $\mathcal{L}^b$. Since the search space of all possible combinations given $b$ elements is $b!$ (factorial), we use an approximate greedy search algorithm limiting the number of combinations to $\frac{(b+1)\cdot b}{2}$. We initialize the set $\mathcal{K}^1$ to contain only the layer whose coefficient $c^l$ is the highest among all layers in $\mathcal{L}^b$. At each subsequent step, the layer $l$, where $l \in \mathcal{L}^b \land l \notin \mathcal{K}^n$, which maximizes the join correlation coefficient $c^{\mathcal{K}^n\cup{\{l\}}}$, is added to the set $\mathcal{K}^n$. To compute $c^{\mathcal{K}^n\cup{\{l\}}}$ we use: % we evaluate individual layers from $\mathcal{L}^b$ not already in the set $\mathcal{K}$. Formally, we evaluate layer $l$, where $l \in \mathcal{L}^b \land l \notin \mathcal{K}$, using:
\begin{equation}\label{equation:spearman_correlation_over_set}
    c^{\mathcal{K}^n\cup{\{l\}}} = \rho(\{\dot{q}_i\}_{i=1}^{N}, \{\hat{q}^{\mathcal{K}^n\cup \{l\}}_i\}_{i=1}^{N}),
\end{equation}
where $\mathcal{K}^n\cup{\{l\}}$ is the union of the set $\mathcal{K}^n$ and the single element set $\{l\}$, and $\{\hat{q}^{\mathcal{K}^n\cup \{l\}}_i\}_{i=1}^{N}$, the set of joint quality scores of the calibration dataset, computed using:
\begin{equation}\label{equation:join_quality_score}
    \{\hat{q}^{\mathcal{K}^n\cup \{l\}}_i\}_{i=1}^{N} = \frac{1}{n + 1} \sum_{j\in \mathcal{K}^n \cup \{l\}}\{\hat{q}^j_i\}_{i=1}^{N},\vspace{-1mm}
\end{equation}
where $n$ is the cardinality of $\mathcal{K}$ in the current step. By performing $b - 1$ steps, we exhaust all elements from $\mathcal{L}^b$ and obtain $b$ possible solutions $\{\mathcal{K}^n\}_{n=1}^{b}$. The $\mathcal{K}^n$, with the highest correlation coefficient $c^{\mathcal{K}^n}$, is selected as the final set of observed representations $\mathcal{K}$. %The final set of observed intermediate layers $\mathcal{K}$ is selected using:

\subsection{Observer Usage}\label{sec:methodology:subsection:observer_usage}

Once the observer has been initialized, i.e., the aggregation function $S(\cdot)$ and the set of representations $\mathcal{K}$ have been determined, for a given FR model $M$, the process of quality assessment is straightforward. During the recognition process, a sample $x$ is run through the FR model $M$. The observer then applies the aggregation function $S(\cdot)$ on the values of the observed representations $z^l_x$, where $l \in \mathcal{K}$, and computes the quality score of the input sample $x$ as:\vspace{0.5mm}
\begin{equation}
    q_x = \frac{1}{|\mathcal{K}|}\sum_{k\in \mathcal{K}} S(z^k_x),\vspace{0.5mm}
\end{equation}
where $|\cdot|$ is the cardinality. Thus, the quality score $q_x$ is the average aggregated value of all observed representations.

%Once the observer has been initialized for a given FR model $M$, meaning that both the aggregation function $S(\cdot)$, which maps any intermediate representation $\{z_i\}_{i=0}^N$ into a single numerical value, and the set $\mathcal{K}$ of indices, which maximizes the biometric utility according to \ref{equation:spearman_correlation_over_set}, its usage is trivial. During the process of face recognition, more specifically the feature extraction step, where an observed face sample $x$ is run through an FR model $M$, the observer will compute the aggregation function for all intermediate layers $k \in \mathcal{K}$, the final quality score 

%\begin{equation}
%    q_x = \frac{1}{|\mathcal{K}|}\sum_{k\in \mathcal{K}} S(z_k)
%\end{equation}

%is computed by calculating the mean value over all responses of the %aggregation function. 

\subsection{Auxiliary FIQA Approach }\label{sec:methodology:subsection:custom_fiqa}

The proposed FROQ technique relies on a calibration set of face images and the corresponding pseudo-quality labels. To extract the labels, an auxiliary FIQA technique is needed. Here, we present a simple new unsupervised FIQA approach for this task. %to distance FROQ from any existing techniques. 
The auxiliary approach is based on three different types of face sample perturbations, i.e., horizontal flipping, Gaussian noising, and partial occlusions. %Previous works have already explored horizontal flips~\cite{faceqan} and Gaussian noise~\cite{diffiqa} for the task of quality assessment, while occlusion, to the best of our knowledge, has not been explored. 

%To distance the presented technique from the success of any previous research, we devise a custom FIQA approach to label the calibration dataset. The approach is based on image perturbations, which have previously been shown to be effective for the task of quality estimation. Given a sample $x$, such methods compare the features of the original $z_x$ and perturbed sample $z_x^P$. To obtain the perturbed sample $x^P$, a perturbation function $P(\cdot)$ is applied on the given sample, while to compare the latent representations, methods most often use the cosine similarity $sim_{cos}$. We propose a new method, which employs three different perturbation functions to the input samples: $(i)$ horizontal flips, $(ii)$ noising and $(iii)$ occlusions. While flips and noising have been used in the past, we are not aware of any previous research regarding sample occlusions.

\begin{figure}[tb!]
    \centering
    \includegraphics[width=0.96\linewidth]{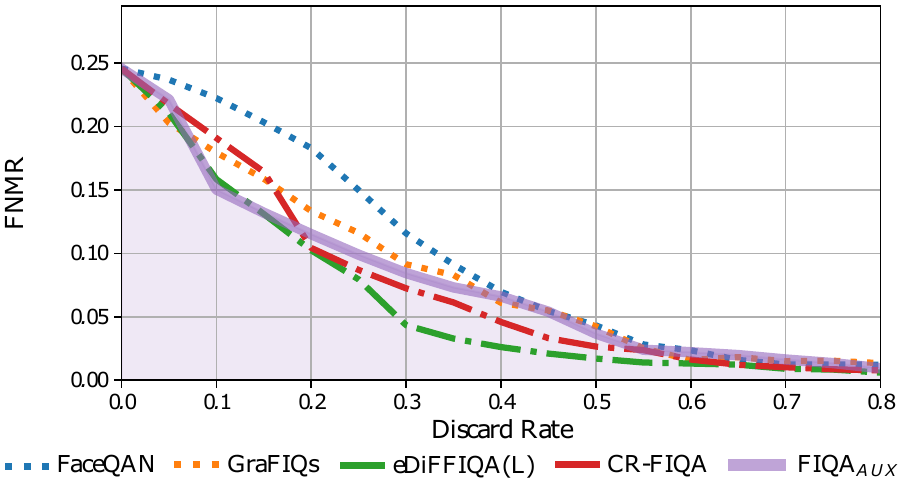}
    \caption{\textbf{Performance of the auxiliary FIQA technique.} 
    We compare the performance of the auxiliary FIQA technique against two unsupervised and two supervised state-of-the-art techniques, on the XQLFW benchmark using the AdaFace FR model.
    }
    \label{fig:edc_auxiliary}
\end{figure}

Given a face sample $x$ from the calibration set, we compute the pseudo-quality label $q_x$ using:
\begin{equation}
    q_x = \frac{1}{3} (q_x^F + q_x^N + q_x^O),
\end{equation}
where $q_x^F$, $q_x^N$, and $q_x^O$ are the partial quality labels obtained using horizontal flips, Gaussian noise, and occlusions. The process to obtain the flip $q_x^F$ and noise $q_x^N$ labels is trivial. Using cosine similarity, we compare the features of the original and a horizontally flipped or noisy sample, respectively. We apply noise to the sample according to:\vspace{-0.5mm}
\begin{equation}
    x^N = (1 - \alpha) \cdot x + \alpha \cdot \mathcal{N}(0, 1),\vspace{-0.5mm}
\end{equation}
where $\alpha$ is a hyperparameter of the approach, and $\mathcal{N}(0, 1)$ is a normally distributed random variable, with a mean of zero and variance of one. For the occlusion perturbation, we compute the cosine similarity between the features of the original $x$ and several occluded samples $\{x^{O_i}\}_{i=1}^{R}$. The sample $x$, which we assume is square ($h$ equals $w$), is first divided into $R=(h/o)^2$ non-overlapping squares, where $h$ is the height of the image and $o$ the desired size of each square. To construct an occluded sample $x^o$, the $o$-th non-overlapping square in the original image is masked by setting all pixels to zero (black). By computing the similarity for all occluded samples and averaging their scores, we obtain the partial quality label $q_x^O$. The performance of the proposed auxiliary FIQA technique is shown in Fig.~\ref{fig:edc_auxiliary}.  % apply several masks individually to the sample $x$, in the process covering each part of the original image. Each mask covers a square part of the image, where no two squares of any two given masks overlap. Formally, given a sample $x$, which we assume is square with both dimensions equal $h$, we construct $(h/o)^2$ masks, where $o^2$ is the number of pixels each mask will occlude. Applying each mask individually we obtain $(h/o)^2$ occluded samples, to combine them into a single score we average the cosine similarities across all obtained occluded samples.

\begin{table}[!t!]
    \centering
    \renewcommand{\arraystretch}{1.1}
    \caption{\textbf{Comparison of benchmarks datasets.} The experiments are performed using eight commonly used benchmarks, which vary in size and focus on different aspects of face image quality.  %Methods are evaluated using eight common benchmarks, each evaluating different quality factors. 
    \vspace{1mm}}%\vspace{-2mm}
\resizebox{\columnwidth}{!}{%    
    \begin{tabular}{l l c c c c c c c c }

        \multirow{ 2}{*}{\textbf{Dataset}} & \multirow{ 2}{*}{\textbf{Images}} & \multirow{ 2}{*}{\textbf{IDs}} & \multicolumn{2}{c}{\textbf{Comparisons}} && \multicolumn{3}{c}{\textbf{Main Focus}} & \multirow{2}{*}{\textbf{Scale}$^\dagger$}\\
        \cline{4-5}\cline{7-9}
        
        & & & Mated & Non-mated && Pose & Age & Quality & \\ 
        \midrule
        LFW~\cite{lfw} & $13{\small,}233$ & $5{\small,}749$ & $3{\small,}000$ & $3{\small,}000$ && \ding{55} & \ding{55} & \ding{55} & M \\ 
        Adience~\cite{adience} & $19{\small,}370$ & $2{\small,}284$ & $20{\small,}000$ & $20{\small,}000$ && \ding{55} & \ding{55} & \ding{55} & M \\
        \dashmidrule  
        CFP-FP~\cite{cfp-fp} & $7{\small,}000$ & $500$ & $3{\small,}500$ & $3{\small,}500$ && \ding{51} & \ding{55} & \ding{55} & M  \\
        CPLFW~\cite{cplfw} & $11{\small,}652$ & $3{\small,}930$& $3{\small,}000$ & $3{\small,}000$ && \ding{51} & \ding{55} & \ding{55} & M \\
        \dashmidrule  
        CALFW~\cite{calfw} & $12{\small,}174$ & $4{\small,}025$ & $3{\small,}000$ & $3{\small,}000$ &&\ding{55} & \ding{51} & \ding{55} & M \\
        AgeDB~\cite{agedb} & $16{\small,}488$ & $570$ & $100{\small,}00$ & $100{\small,}00$ && \ding{55} & \ding{51} & \ding{55} & M \\
        \dashmidrule  
        XQLFW~\cite{xqlfw} & $13{\small,}233$ & $5{\small,}749$ & $3{\small,}000$ & $3{\small,}000$ && \ding{55} & \ding{55} & \ding{51} & M\\
        \dashmidrule  
        IJB-C~\cite{ijbc} & $23{\small,}124^{\ddagger}$ & $3{\small,}531$ & $19{\small,}557$ & $15{\small,}638{\small,}932$ && \ding{55} & \ding{55} & \ding{55} & L\\

        \bottomrule
        %\multicolumn{10}{l}{\footnotesize $^\dagger$O-E - Occlusion, Expression; B-R-N - Blur, Resolution, Noise; Sc - Scale.}\\
        \multicolumn{8}{l}{\footnotesize $^\dagger$M - Medium; L - Large; Values estimated subjectively by the authors.}\\
        \multicolumn{8}{l}{\footnotesize $^{\ddagger}$ number of templates, each containing several images}
        \vspace{-6mm}
    \end{tabular}
    }
    \label{tab:dataset_info}
\end{table}

\subsection{Calibration Dataset}\label{sec:methodology:subsection:calibration_dataset}

Discovering informative intermediate representations of a given FR model, as described in Sec.~\ref{sec:methodology:subsection:observer_definition}, requires a small calibration set of face images. To guarantee a fair experimental evaluation, the set of images should not overlap with any of the benchmarks used in the experiments. Therefore, we construct our calibration set from the popular large-scale Glint360K\cite{glint1, glint2} dataset. To create our subset, we randomly selected $500$ samples from the original dataset. Since the images in the original dataset are mostly high- to medium-quality, we further degrade $33\%$ of the images using the BSRGAN~\cite{bsrgan} degradation model. This means that the observer initialization step focuses on a wide range of image qualities when constructing the observed set of intermediate representations, ensuring good overall performance of the final quality observer.

\begin{table}[t]
  \centering
  \caption{\textbf{Comparison of FIQA techniques.} We analyze ten methods, and compare their requirements pre- and during-inference to the proposed FROQ technique. The unsupervised methods are marked using \textcolor{blue!60}{BLUE}, and the supervised using \textcolor{green!60}{GREEN} stripes. % The evaluated methods exhibit variations in requirements and inference conditions. We mark the unsupervised methods using \textcolor{blue!40}{BLUE}, and supervised using \textcolor{green!40}{GREEN} stripes.
  }
  \resizebox{\columnwidth}{!}{%
    \begin{NiceTabular}{c l | *{4}{c} c c c c c}
        %{pattern = north west lines,\\ pattern color = blue}
      % Header rows
         & \multicolumn{5}{c}{~} & \multicolumn{5}{c}{\textbf{Inference}} \\
             \cline{7-11}
        & \textbf{Method} & \textbf{\rotatebox{90}{\makecell[{{l}}]{\noindent Quality\\Labels}}} & \textbf{\rotatebox{90}{\makecell[{{l}}]{\noindent Architecture\\Specific}}} & \textbf{\rotatebox{90}{\makecell[{{l}}]{\noindent Additional\\Training}}} & \textbf{\rotatebox{90}{\makecell[{{l}}]{Custom\\Loss}}} 
        & \textbf{\rotatebox{90}{\makecell[{{l}}]{\noindent Feed-\\Forward}}} 
        & \textbf{\rotatebox{90}{\makecell[{{l}}]{\noindent Backwards}}}
        & \textbf{\rotatebox{90}{\makecell[{{l}}]{\noindent Feature\\Level}}}
        & \textbf{\rotatebox{90}{\makecell[{{l}}]{\noindent Gradient\\Level}}}
         & \textbf{\rotatebox{90}{\makecell[{{l}}]{\noindent Representation\\Level}}}\\
        \midrule
      % Group A rows
      \Block[tikz={pattern = {Lines[angle=-45, distance=1.0mm,  line width=0.5mm]},pattern color=blue!40}]{4-1} 
      ~ & SER-FIQ~\cite{serfiq}       & \ding{55} & \ding{51} & \ding{55} & \ding{55} & 100 & 0   & \ding{51} & \ding{55} & \ding{55} \\
      ~ & FaceQAN~\cite{faceqan}      & \ding{55} & \ding{55} & \ding{55} & \ding{55} & 10  & 10  & \ding{51} & \ding{51} & \ding{55} \\
      ~ & GraFIQs~\cite{grafiqs}      & \ding{55} & \ding{55} & \ding{55} & \ding{55} & 1   & 1   & \ding{55} & \ding{51} & \ding{55} \\
      \midrule
      % Group B rows
      \Block[tikz={pattern = {Lines[angle=-45, distance=1.0mm,  line width=0.5mm]},pattern color=green!40
      }]{8-1} ~ & SDD-FIQA~\cite{sdd-fiqa}     & \ding{51} & \ding{55} & \ding{51} & \ding{55} & 1   & 0   & \ding{51} & \ding{55} & \ding{55} \\
      {} & LightQnet~\cite{lightqnet}& \ding{51} & \ding{55} & \ding{51} & \ding{51} & 1   & 0   & \ding{51} & \ding{55} & \ding{55} \\
      {} & PCNet~\cite{pcnet}           & \ding{51} & \ding{55} & \ding{51} & \ding{55} & 1   & 0   & \ding{51} & \ding{55} & \ding{55} \\
      {} & eDiFFIQA(L)~\cite{ediffiqa}& \ding{51} & \ding{55} & \ding{51} & \ding{51} & 1   & 0   & \ding{51} & \ding{55} & \ding{55} \\
      {} & CLIB-FIQA~\cite{clib-fiqa}  & \ding{51} & \ding{51} & \ding{51} & \ding{51} & 1   & 0   & \ding{51} & \ding{55} & \ding{55} \\
      \dashmidrule
      {} & MagFace~\cite{magface}      & \ding{55} & \ding{55} & \ding{51} & \ding{51} & 1   & 0   & \ding{51} & \ding{55} & \ding{55} \\
      {} & CR-FIQA~\cite{cr-fiqa}      & \ding{55} & \ding{55} & \ding{51} & \ding{51} & 1   & 0   & \ding{51} & \ding{55} & \ding{55} \\
      \midrule
      % Last method with dual stripes
      \rowcolor{gray!10}
      \Block[tikz={preaction={fill, blue!40}, pattern = {Lines[angle=-45, distance=1.0mm,  line width=0.5mm]},pattern color=green!40}]{1-1} {} & FROQ   & \ding{51} & \ding{55} & \ding{55} & \ding{55} & 1   & 0   & \ding{55} & \ding{55} & \ding{51} \\
      \bottomrule
    \end{NiceTabular}%
  }\vspace{-3mm}
  \label{tab:methods_info}
\end{table}

\begin{figure*}[ht!]
    \centering
    \includegraphics[width=0.83\linewidth]{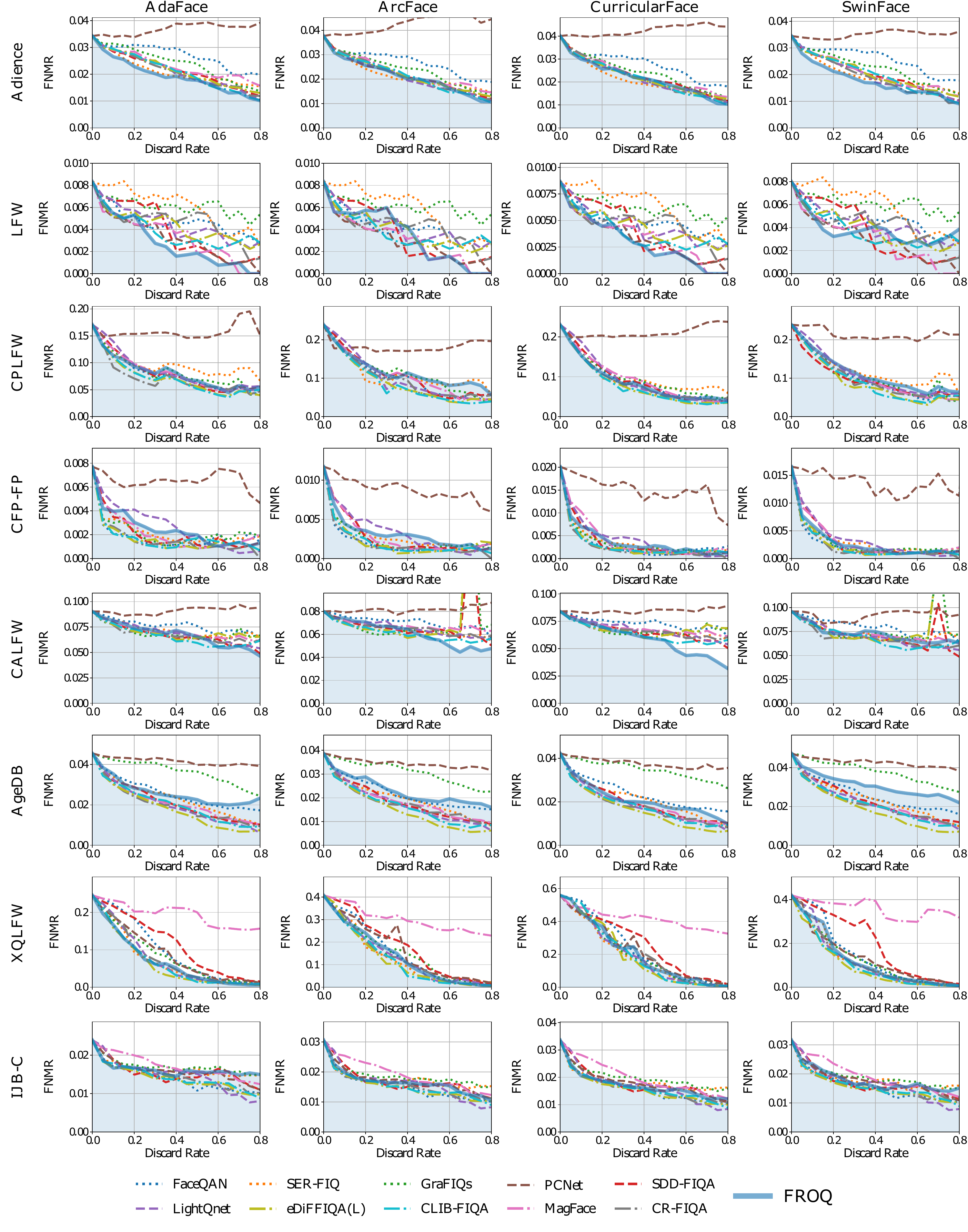}\vspace{-2mm}
    \caption{\textbf{Comparison to the state-of-the-art using EDC curves.} 
    The performance of FROQ is compared against ten recent FIQA techniques, on eight benchmark datasets, using four FR models. The curves present FNMR values at FMR=$1e^{-3}$, for different discard rates. The unsupervised methods use dotted lines, the supervised methods dashed or dashed-dotted lines, and our method a full line.
    }
    \label{fig:edc_curves}\vspace{-2mm}
\end{figure*}

\section{Experiments \& Results}\label{sec:experiments_and_results}

\subsection{Experimental Setup}\label{sec:experiments_and_results:subsection:experimental_setup}

\noindent \textbf{Experimental Setting.}~We compare the performance of FROQ to $10$ state-of-the-art FIQA techniques i.e.: $(i)$ the unsupervised SER-FIQ~\cite{serfiq}, FaceQAN~\cite{faceqan}, GraFIQs~\cite{grafiqs} methods, and $(ii)$ the supervised PCNet~\cite{pcnet}, SDD-FIQA~\cite{sdd-fiqa}, LightQnet~\cite{lightqnet}, eDifFIQA~\cite{ediffiqa}, CLIB-FIQA~\cite{clib-fiqa}, MagFace~\cite{magface} and CR-FIQA~\cite{cr-fiqa} methods. All techniques are summarized in Table~\ref{tab:methods_info}. We perform experiments using $4$ commonly used state-of-the-art FR models, showing that our method can perform well using any modern FR model.~Specifically we use the CNN-based: AdaFace\footnote{\label{fnote:adaface}\scriptsize{\url{https://github.com/mk-minchul/AdaFace}}}~\cite{adaface}, ArcFace\footnote{\label{fnote:arcface}\scriptsize{\url{https://github.com/deepinsight/insightface}}}~\cite{arcface}, and CurricularFace\footnote{\label{fnote:curricularface}\scriptsize{\url{https://github.com/HuangYG123/CurricularFace}}}~\cite{curricularface} models, as well as the Transformer based SwinFace\footnote{\label{fnote:swinface}\scriptsize{\url{https://github.com/lxq1000/SwinFace}}}~\cite{swinface} model.~The CNN models all use the ResNet100 backbone, while SwinFace uses the Swin Transformer~\cite{swin}. All models are trained on the WebFace12M$^{\ref{fnote:adaface}}$, MS1MV3$^{~\ref{fnote:arcface},\ref{fnote:swinface}}$, Glint360k$^{~\ref{fnote:arcface}}$, and CASIA-WebFace$^{~\ref{fnote:curricularface}}$ datasets.~To show that FROQ can perform well in different scenarios, we repeat the experiments on $8$ different face benchmarks, summarized in Table~\ref{tab:dataset_info}, i.e.: $(i)$ LFW~\cite{lfw}, Adience~\cite{adience}, $(ii)$ cross-pose CPLFW~\cite{cplfw}, CFP-FP~\cite{cfp-fp}, $(iii)$ cross-age CALFW~\cite{calfw}, AgeDB~\cite{agedb}, $(iv)$ cross-quality XQLFW~\cite{xqlfw} and $(v)$ the large-scale IJB-C~\cite{ijbc} benchmark.

\vspace{1.1mm}\noindent\textbf{Evaluation Methodology.} To evaluate the methods, we use EDC (Error-versus Discard Characteristic) curves and the pAUC (partial Area Under the Curve) values calculated using the EDC curves. Both of these are regularly used to evaluate FIQA techniques \cite{cr-fiqa,ediffiqa,survey,magface}. The EDC curves measure the FNMR (False Non-Match Rate) at a given FMR (False Match Rate), typically $1e-3$, for different discard rates of low-quality images. Overall, the EDC curves show how the performance of a given FR model improves when discarding some percentage of the lowest quality images from the dataset. The pAUC values condense the information shown by the EDC curves into a single numerical score by calculating the area under the curves up to some percentage of discarded images, typically set to $20\%$. For a more concise and clear presentation, we normalize the pAUC values using the value calculated at $0\%$ discard rate.

\vspace{1.5mm}\noindent\textbf{Implementation Details.}~In the \textit{Observer Initialization} step, we use the top $b$ individual layers to construct the final set $\mathcal{K}$. To balance the complexity of the search algorithm and the completeness of the final set, we use $b=10$. % When calculating the final set $\mathcal{K}$, we set $b$ to $10$. The parameter $m$ determines how many of the top-scoring layers from a model we will consider when building the final set of observed intermediate layers $\mathcal{K}$. In this sense, $m$ balances the size of the search space and the completeness of the final set. 
The proposed auxiliary FIQA technique uses three separate perturbations to produce the pseudo quality scores $\dot{q}$. For the noisy component $q^N$, we use the hyperparameter $\alpha$, while for the occlusion component $q^O$, we use the hyperparameter $o$. The parameter $\alpha$ determines the amount of noise added to each sample. For the sake of quality assessment, the amount should be minimal, therefore, we set $\alpha = 0.001$. On the other hand, $o$ determines the size of the square-structured occlusions applied to the sample, measured in pixels. For modern FR models, images are usually resized to $112\times 112$ pixels, and we, therefore, choose $o=14$, to divide the image into non-overlapping squares. All experiments are conducted on a PC with an Intel i9-10900KF CPU, $64$ GB of RAM, and an Nvidia $3090$ GPU. Observer initialization for a given FR model takes less than five minutes on the presented PC. For ArcFace, CurricularFace, and SwinFace, the process selects four and for the AdaFace model, five intermediate representation to be observed during recognition. 

% Parameters of method: just m?
% Parameters of FIQA: alpha, m

\begin{table}[!ht]
    \centering
    \caption{\textbf{Comparison to the state-of-the-art.} 
    The table shows pAUC scores calculated at a discard rate of $0.2$, with FMR set to $10^{-3}$, over all benchmarks using all four FR models. We mark the \textbf{best} and \textit{second best} result of each dataset. The last column shows average results over all benchmarks, the best average result of the unsupervised methods is colored \begin{tabular}{c}\cellcolor{blue!10}BLUE\end{tabular} and the best of supervised methods is colored \begin{tabular}{c}\cellcolor{green!10}GREEN\end{tabular}. Unsupervised methods are marked using \textcolor{blue!60}{BLUE}, and supervised using \textcolor{green!60}{GREEN} stripes.
    %The table reports pAUC scores at a discard rate of $0.3$ and a FMR of $10^{-3}$. Average results across all datasets are marked $\overline{\text{pAUC}}$. The best result for each dataset is shown in bold, the overall best result is colored green, the second best blue and the third best red.
    }
    \resizebox{\columnwidth}{!}{%
\begin{NiceTabular}{c l | l l l l l l l l | r}%[hvlines]

 \Block{2-11}{\textbf{AdaFace\cite{adaface}} - $pAUC($FMR$=1e^{-3})[\downarrow]$} \\
 \\
 {} & \textbf{{Methods}} & \textbf{Adience} & \textbf{LFW} & \textbf{CPLFW} & \textbf{CFP-FP} & \textbf{CALFW} & \textbf{AgeDB} & \textbf{XQLFW} & \textbf{IJB-C} & $\overline{pAUC}$ \\
\midrule\Block[tikz={pattern = {Lines[angle=-45, distance=1.0mm,  line width=0.5mm]},pattern color=blue!40}]{4-1} {} & \textbf{SER-FIQ\cite{serfiq}} & $\mathit{0.871}$ & $0.982$ & $0.775$ & $0.563$ & $0.930$ & $0.809$ & $0.809$ & $0.812$ & $0.819$ \\
{} & \textbf{FaceQAN\cite{faceqan}} & $0.919$ & $0.797$ & $0.808$ & $\mathbf{0.474}$ & $0.945$ & $0.833$ & $0.924$ & $0.800$ & \cellcolor{blue!10}$0.813$ \\
{} & \textbf{GraFIQs\cite{grafiqs}} & $0.932$ & $0.863$ & $0.791$ & $0.557$ & $\mathit{0.893}$ & $0.950$ & $0.801$ & $0.846$ & $0.829$ \\
\dashmidrule
\Block[tikz={pattern = {Lines[angle=-45, distance=1.0mm,  line width=0.5mm]},pattern color=green!40}]{8-1} {} & \textbf{PCNet\cite{pcnet}} & $1.003$ & $\mathbf{0.730}$ & $0.914$ & $0.893$ & $0.985$ & $0.969$ & $0.826$ & $0.843$ & $0.895$ \\
{} & \textbf{SDD-FIQA\cite{sdd-fiqa}} & $0.884$ & $0.857$ & $0.819$ & $0.632$ & $0.911$ & $0.796$ & $0.907$ & $0.854$ & $0.832$ \\
{} & \textbf{LightQnet\cite{lightqnet}} & $0.890$ & $0.837$ & $0.854$ & $0.711$ & $0.925$ & $0.792$ & $0.835$ & $0.846$ & $0.836$ \\
{} & \textbf{eDiFFIQA(L)\cite{ediffiqa}} & $0.889$ & $0.751$ & $\mathit{0.736}$ & $0.490$ & $0.894$ & $\mathbf{0.756}$ & $\mathit{0.759}$ & $\mathbf{0.791}$ & \cellcolor{green!10}$0.758$ \\
{} & \textbf{CLIB-FIQA\cite{clib-fiqa}} & $0.893$ & $0.762$ & $0.746$ & $\mathit{0.479}$ & $0.897$ & $0.770$ & $0.770$ & $0.807$ & $0.766$ \\
{} & \textbf{MagFace\cite{magface}} & $0.890$ & $\mathit{0.735}$ & $0.805$ & $0.632$ & $0.900$ & $\mathit{0.760}$ & $0.958$ & $0.915$ & $0.824$ \\
{} & \textbf{CR-FIQA\cite{cr-fiqa}} & $0.890$ & $0.755$ & $\mathbf{0.699}$ & $0.504$ & $\mathbf{0.887}$ & $0.763$ & $0.833$ & $\mathit{0.796}$ & $0.766$ \\
\midrule
\rowcolor{gray!10}
\Block[tikz={preaction={fill, blue!40}, pattern = {Lines[angle=-45, distance=1.0mm,  line width=0.5mm]},pattern color=green!40}]{1-1}{} & \textbf{FROQ$_{ADA}$} & $\mathbf{0.843}$ & $0.754$ & $0.764$ & $0.646$ & $0.925$ & $0.822$ & $\mathbf{0.753}$ & $0.798$ & $\underline{0.788}$ \\
\bottomrule

 \Block{2-11}{\textbf{ArcFace\cite{arcface}} - $pAUC($FMR$=1e^{-3})[\downarrow]$} \\
 \\
 {} & \textbf{{Methods}} & \textbf{Adience} & \textbf{LFW} & \textbf{CPLFW} & \textbf{CFP-FP} & \textbf{CALFW} & \textbf{AgeDB} & \textbf{XQLFW} & \textbf{IJB-C} & $\overline{pAUC}$ \\
\midrule\Block[tikz={pattern = {Lines[angle=-45, distance=1.0mm,  line width=0.5mm]},pattern color=blue!40}]{4-1} {} & \textbf{SER-FIQ\cite{serfiq}} & $\mathit{0.840}$ & $0.982$ & $0.797$ & $0.539$ & $0.934$ & $0.790$ & $0.828$ & $0.732$ & $0.805$ \\
{} & \textbf{FaceQAN\cite{faceqan}} & $0.864$ & $0.775$ & $0.826$ & $\mathbf{0.457}$ & $0.962$ & $0.835$ & $0.883$ & $0.731$ & \cellcolor{blue!10}$0.792$ \\
{} & \textbf{GraFIQs\cite{grafiqs}} & $0.872$ & $0.863$ & $0.814$ & $0.538$ & $\mathit{0.902}$ & $0.946$ & $0.818$ & $0.786$ & $0.817$ \\
\dashmidrule
\Block[tikz={pattern = {Lines[angle=-45, distance=1.0mm,  line width=0.5mm]},pattern color=green!40}]{8-1} {} & \textbf{PCNet\cite{pcnet}} & $1.012$ & $\mathbf{0.697}$ & $0.810$ & $0.920$ & $0.998$ & $0.965$ & $0.860$ & $0.770$ & $0.879$ \\
{} & \textbf{SDD-FIQA\cite{sdd-fiqa}} & $0.841$ & $0.857$ & $0.829$ & $0.649$ & $0.931$ & $0.801$ & $0.935$ & $0.806$ & $0.831$ \\
{} & \textbf{LightQnet\cite{lightqnet}} & $0.840$ & $0.814$ & $0.862$ & $0.657$ & $0.930$ & $0.797$ & $0.824$ & $0.788$ & $0.814$ \\
{} & \textbf{eDiFFIQA(L)\cite{ediffiqa}} & $0.842$ & $0.751$ & $\mathbf{0.771}$ & $0.497$ & $0.904$ & $\mathbf{0.757}$ & $0.810$ & $\mathit{0.729}$ & \cellcolor{green!10}$0.758$ \\
{} & \textbf{CLIB-FIQA\cite{clib-fiqa}} & $0.846$ & $0.762$ & $\mathit{0.778}$ & $0.502$ & $\mathbf{0.900}$ & $\mathit{0.762}$ & $\mathbf{0.789}$ & $0.730$ & $0.759$ \\
{} & \textbf{MagFace\cite{magface}} & $0.852$ & $\mathit{0.712}$ & $0.809$ & $0.634$ & $0.925$ & $0.771$ & $0.960$ & $0.867$ & $0.816$ \\
{} & \textbf{CR-FIQA\cite{cr-fiqa}} & $0.861$ & $0.732$ & $0.791$ & $\mathit{0.477}$ & $0.912$ & $0.764$ & $0.814$ & $\mathbf{0.724}$ & $0.759$ \\
\midrule
\rowcolor{gray!10}
\Block[tikz={preaction={fill, blue!40}, pattern = {Lines[angle=-45, distance=1.0mm,  line width=0.5mm]},pattern color=green!40}]{1-1}{} & \textbf{FROQ$_{ARC}$} & $\mathbf{0.827}$ & $0.746$ & $0.796$ & $0.571$ & $0.937$ & $0.837$ & $\mathit{0.805}$ & $0.745$ & $\underline{0.783}$ \\
\bottomrule

 \Block{2-11}{\textbf{CurricularFace\cite{curricularface}} - $pAUC($FMR$=1e^{-3})[\downarrow]$} \\
 \\
 {} & \textbf{{Methods}} & \textbf{Adience} & \textbf{LFW} & \textbf{CPLFW} & \textbf{CFP-FP} & \textbf{CALFW} & \textbf{AgeDB} & \textbf{XQLFW} & \textbf{IJB-C} & $\overline{pAUC}$ \\
\midrule\Block[tikz={pattern = {Lines[angle=-45, distance=1.0mm,  line width=0.5mm]},pattern color=blue!40}]{4-1} {} & \textbf{SER-FIQ\cite{serfiq}} & $0.832$ & $0.986$ & $0.764$ & $0.493$ & $0.926$ & $0.794$ & $\mathbf{0.840}$ & $\mathit{0.725}$ & $0.795$ \\
{} & \textbf{FaceQAN\cite{faceqan}} & $0.855$ & $0.786$ & $0.804$ & $0.453$ & $0.953$ & $0.830$ & $0.931$ & $0.730$ & \cellcolor{blue!10}$0.793$ \\
{} & \textbf{GraFIQs\cite{grafiqs}} & $0.857$ & $0.882$ & $0.785$ & $0.477$ & $\mathit{0.906}$ & $0.950$ & $0.887$ & $0.780$ & $0.815$ \\
\dashmidrule
\Block[tikz={pattern = {Lines[angle=-45, distance=1.0mm,  line width=0.5mm]},pattern color=green!40}]{8-1} {} & \textbf{PCNet\cite{pcnet}} & $1.000$ & $\mathbf{0.732}$ & $0.902$ & $0.931$ & $0.993$ & $0.969$ & $\mathit{0.855}$ & $0.776$ & $0.895$ \\
{} & \textbf{SDD-FIQA\cite{sdd-fiqa}} & $0.838$ & $0.865$ & $0.812$ & $0.556$ & $0.932$ & $0.793$ & $0.867$ & $0.806$ & $0.809$ \\
{} & \textbf{LightQnet\cite{lightqnet}} & $\mathit{0.827}$ & $0.834$ & $0.852$ & $0.574$ & $0.938$ & $0.783$ & $0.855$ & $0.787$ & $0.806$ \\
{} & \textbf{eDiFFIQA(L)\cite{ediffiqa}} & $0.831$ & $0.763$ & $\mathbf{0.751}$ & $0.448$ & $0.906$ & $\mathbf{0.740}$ & $0.883$ & $\mathbf{0.721}$ & \cellcolor{green!10}$0.755$ \\
{} & \textbf{CLIB-FIQA\cite{clib-fiqa}} & $0.834$ & $0.774$ & $\mathit{0.756}$ & $\mathit{0.446}$ & $\mathbf{0.905}$ & $\mathit{0.749}$ & $0.910$ & $0.733$ & $0.763$ \\
{} & \textbf{MagFace\cite{magface}} & $0.841$ & $\mathit{0.736}$ & $0.792$ & $0.624$ & $0.921$ & $0.757$ & $0.901$ & $0.875$ & $0.806$ \\
{} & \textbf{CR-FIQA\cite{cr-fiqa}} & $0.859$ & $0.746$ & $0.765$ & $\mathbf{0.428}$ & $0.908$ & $0.751$ & $0.901$ & $0.734$ & $0.761$ \\
\midrule
\rowcolor{gray!10}
\Block[tikz={preaction={fill, blue!40}, pattern = {Lines[angle=-45, distance=1.0mm,  line width=0.5mm]},pattern color=green!40}]{1-1}{} & \textbf{FROQ$_{CURR}$} & $\mathbf{0.821}$ & $0.757$ & $0.757$ & $0.531$ & $0.922$ & $0.795$ & $0.873$ & $0.735$ & $\underline{0.774}$ \\
\bottomrule

 \Block{2-11}{\textbf{SwinFace\cite{swinface}} - $pAUC($FMR$=1e^{-3})[\downarrow]$} \\
 \\
{} & \textbf{{Methods}} & \textbf{Adience} & \textbf{LFW} & \textbf{CPLFW} & \textbf{CFP-FP} & \textbf{CALFW} & \textbf{AgeDB} & \textbf{XQLFW} & \textbf{IJB-C} & $\overline{pAUC}$ \\
\midrule\Block[tikz={pattern = {Lines[angle=-45, distance=1.0mm,  line width=0.5mm]},pattern color=blue!40}]{4-1} {} & \textbf{SER-FIQ\cite{serfiq}} & $\mathit{0.840}$ & $1.002$ & $0.801$ & $0.496$ & $0.924$ & $0.799$ & $0.824$ & $0.746$ & $0.804$ \\
{} & \textbf{FaceQAN\cite{faceqan}} & $0.896$ & $0.796$ & $0.820$ & $\mathbf{0.441}$ & $0.949$ & $0.827$ & $0.934$ & $0.759$ &\cellcolor{blue!10}$0.803$ \\
{} & \textbf{GraFIQs\cite{grafiqs}} & $0.886$ & $0.889$ & $0.819$ & $0.500$ & $\mathit{0.889}$ & $0.952$ & $0.791$ & $0.798$ & $0.815$ \\
\dashmidrule
\Block[tikz={pattern = {Lines[angle=-45, distance=1.0mm,  line width=0.5mm]},pattern color=green!40}]{8-1} {} & \textbf{PCNet\cite{pcnet}} & $0.980$ & $\mathbf{0.716}$ & $0.996$ & $0.972$ & $0.962$ & $0.963$ & $0.766$ & $0.789$ & $0.893$ \\
{} & \textbf{SDD-FIQA\cite{sdd-fiqa}} & $0.859$ & $0.871$ & $\mathbf{0.738}$ & $0.604$ & $0.908$ & $0.811$ & $0.922$ & $0.816$ & $0.816$ \\
{} & \textbf{LightQnet\cite{lightqnet}} & $0.867$ & $0.837$ & $0.863$ & $0.603$ & $0.914$ & $0.798$ & $0.815$ & $0.799$ & $0.812$ \\
{} & \textbf{eDiFFIQA(L)\cite{ediffiqa}} & $0.856$ & $0.771$ & $\mathit{0.784}$ & $0.482$ & $0.897$ & $\mathbf{0.753}$ & $\mathbf{0.715}$ & $\mathbf{0.743}$ & \cellcolor{green!10}$0.750$ \\
{} & \textbf{CLIB-FIQA\cite{clib-fiqa}} & $0.861$ & $0.784$ & $0.791$ & $0.494$ & $0.891$ & $0.765$ & $\mathit{0.741}$ & $0.749$ & $0.759$ \\
{} & \textbf{MagFace\cite{magface}} & $0.862$ & $0.732$ & $0.819$ & $0.591$ & $0.894$ & $\mathit{0.758}$ & $0.960$ & $0.881$ & $0.812$ \\
{} & \textbf{CR-FIQA\cite{cr-fiqa}} & $0.855$ & $0.753$ & $0.801$ & $\mathit{0.466}$ & $\mathbf{0.863}$ & $0.766$ & $0.788$ & $\mathit{0.743}$ & $0.754$ \\
\midrule
\rowcolor{gray!10}
\Block[tikz={preaction={fill, blue!40}, pattern = {Lines[angle=-45, distance=1.0mm,  line width=0.5mm]},pattern color=green!40}]{1-1}{} & \textbf{FROQ$_{SWIN}$} & $\mathbf{0.802}$ & $\mathit{0.726}$ & $0.823$ & $0.525$ & $0.899$ & $0.856$ & $0.828$ & $0.795$ & $\underline{0.782}$ \\
\bottomrule

\end{NiceTabular}
}\vspace{-4mm}
        \label{tab:pauc_values}
\end{table}
%

\iffalse
"""
froq : 0.01102523199061386 \pm 0.0012613854606453743

ser-fiq : 0.11837566728576784 \pm 0.02934077164212967
faceqan : 0.352123 \pm 0.013515316
grafiqs : 0.055698913646142496 \pm 0.0323287240243779

pcnet : 0.013913906882663779 \pm 0.005542158390328578
sdd-fiqa : 0.0050606234835936965 \pm 0.0013002329167619354
lightqnet : 0.004929548991579574 \pm 0.004615436437572028
ediffiqal : 0.010062044226987232 \pm 0.0013425500572474124
clib-fiqa : 0.08034667717953761 \pm 0.05312278654529778

magface : 0.0082195637103117522 \pm 0.00022835872398118182
cr-fiqa : 0.009381163589884432 \pm 0.0003097197140017786

"""
\fi

\begin{table*}[t]
    \centering
    \caption{\textbf{Runtime Complexity of FIQA techniques.} The table shows the method's inference runtime (in $ms$), for a single image, calculated over a set of $10.000$ images.  Unsupervised methods are marked using \textcolor{blue!60}{BLUE}, and supervised using \textcolor{green!60}{GREEN} stripes. \vspace{1mm}}

    \resizebox{\textwidth}{!}{%
    \begin{NiceTabular}{ l | ccc | ccccc |[tikz={dashed}] cc | c}
        %\toprule

        \multirow{2}{*}{{\textbf{FIQA Model}}} & \Block[tikz={pattern = {Lines[angle=-45, distance=1.0mm,  line width=0.5mm]},pattern color=blue!40}]{1-3}{} & {} & {} &
        \Block[tikz={pattern = {Lines[angle=-45, distance=1.0mm,  line width=0.5mm]},pattern color=green!40}]{1-7}{} & {} & {} & {} & {} & {} & {} &
        \Block[tikz={preaction={fill, blue!40}, pattern = {Lines[angle=-45, distance=1.0mm,  line width=0.5mm]},pattern color=green!40}]{1-1}{} \\
        {} & {\textbf{SER-FIQ} \cite{serfiq}} &{\textbf{FaceQAN} \cite{faceqan}} & {\textbf{GraFIQs} \cite{grafiqs}} & {\textbf{PCNet} \cite{pcnet}} & {\textbf{SDD-FIQA} \cite{sdd-fiqa}} & {\textbf{LightQnet} \cite{lightqnet}} & {\textbf{eDifFIQA(L)} \cite{ediffiqa}} & {\textbf{CLIB-FIQA} \cite{clib-fiqa}} & {\textbf{MagFace} \cite{magface}} & {\textbf{CR-FIQA} \cite{cr-fiqa}} & \cellcolor{gray!10}{\textbf{FROQ}} \\
       \midrule
       Runtime ($\mu \pm\sigma$)  & $118.376\pm29.240$  & $352.123\pm13.515$  & $55.698\pm32.328$  & $13.913\pm5.542$  & $5.060\pm1.300$ & $4.929\pm4.615$  & $10.062\pm1.342$  & $80.346\pm53.122$  & $8.219\pm0.228$  & $9.381\pm0.309$   & \cellcolor{gray!10}$11.025\pm1.261$ \\

        \bottomrule
    \end{NiceTabular}
    }
    \vspace{-3mm}
    \label{tab:time_complexity}
\end{table*}

\begin{table*}[t]
    \centering
    \caption{\textbf{Runtime Complexity of FIQA techniques.} The table shows the method's inference runtime (in $ms$), for a single image, calculated over a set of $10.000$ images.  Unsupervised methods are marked using \textcolor{blue!60}{BLUE}, and supervised using \textcolor{green!60}{GREEN} stripes. \vspace{1mm}}

    \resizebox{\textwidth}{!}{%
    \begin{NiceTabular}{ l | ccc | ccccc |[tikz={dashed}] cc | c}
        %\toprule

        \multirow{2}{*}{{\textbf{FIQA Model}}} & \Block[tikz={pattern = {Lines[angle=-45, distance=1.0mm,  line width=0.5mm]},pattern color=blue!40}]{1-3}{} & {} & {} &
        \Block[tikz={pattern = {Lines[angle=-45, distance=1.0mm,  line width=0.5mm]},pattern color=green!40}]{1-7}{} & {} & {} & {} & {} & {} & {} &
        \Block[tikz={preaction={fill, blue!40}, pattern = {Lines[angle=-45, distance=1.0mm,  line width=0.5mm]},pattern color=green!40}]{1-1}{} \\
        {} & {\textbf{SER-FIQ} \cite{serfiq}} &{\textbf{FaceQAN} \cite{faceqan}} & {\textbf{GraFIQs} \cite{grafiqs}} & {\textbf{PCNet} \cite{pcnet}} & {\textbf{SDD-FIQA} \cite{sdd-fiqa}} & {\textbf{LightQnet} \cite{lightqnet}} & {\textbf{eDifFIQA(L)} \cite{ediffiqa}} & {\textbf{CLIB-FIQA} \cite{clib-fiqa}} & {\textbf{MagFace} \cite{magface}} & {\textbf{CR-FIQA} \cite{cr-fiqa}} & \cellcolor{gray!10}{\textbf{FROQ}} \\
       \midrule
       $\overline{pAUC}$  & $0.819$ & \cellcolor{blue!10}$0.813$ & $0.829$ & $0.895$ & $0.832$ & $0.836$ & \cellcolor{green!10}$0.758$ & $0.766$ & $0.824$ & $0.766$ & \cellcolor{gray!10}$0.788$ \\

        \bottomrule
    \end{NiceTabular}
    }
    \vspace{-3mm}
    \label{tab:time_complexity}
\end{table*}

\subsection{Comparison with the State-of-the-Art}\label{sec:experiments_and_results:subsection:results}

In this section, we compare the experimental results of FROQ to other state-of-the-art methods, by looking at their: $(i)$ performance, and $(ii)$ inference runtime.

\vspace{1.5mm}\noindent\textbf{Performance Analysis.}  In Fig. \ref{fig:edc_curves}, we show the comparison with state-of-the-art methods using EDC curves, and in Table \ref{tab:pauc_values} the corresponding comparison using the pAUC values. Following \cite{faceqan, diffiqa, ediffiqa, clib-fiqa, busch_ifihaveto}, we use a discard rate of $20\%$ to compute the pAUC values. The results are presented for all four FR models and all included benchmark datasets. Observing the results, we see that the proposed FROQ technique performs well on all benchmark datasets and FR models. % FROQ surpasses all unsupervised state-of-the-art methods, and only lags behind three supervised methods: eDifFIQA(L), CR-FIQA, and CLIB-FIQA. Unlike FROQ, all three methods require substantial computational resources to train their respective quality assessment model. 
Looking at individual results, FROQ outperforms all methods on the Adience benchmark regardless of the given FR model. It achieves excellent (second-best) results on the XQLFW and the LFW benchmarks, using AdaFace, ArcFace, and SwinFace models, respectively. While on the CFP-FP and AgeDB benchmarks, FROQ is competitive compared to the unsupervised methods, and slightly behind the average of the state-of-the-art supervised methods. By observing the $\overline{pAUC}$ results, which combine the results of all benchmark datasets, we see that FROQ outperforms the best unsupervised method, FaceQAN, in all scenarios. The proposed method performs worse than the three best supervised methods: eDifFIQA(L), CR-FIQA, and CLIB-FIQA. Compared to FROQ, all three supervised methods require substantial computational resources and additional parameters to train their respective quality assessment models.

\vspace{1.5mm}\noindent\textbf{Runtime Analysis.} In Table \ref{tab:time_complexity} we show the comparison with state-of-the-art methods in terms of the inference runtime. The results measure the mean $\mu$ and standard deviation $\sigma$ of the inference runtime, for a single image, computed over a set of $10.000$ images. To ensure a fair comparison, each image was processed individually (batch size of $1$), the experiments were conducted on the same hardware, and the official implementations provided by the authors were used for all methods. %The final results present a rough estimation of the runtime complexity of individual techniques. 
The proposed FROQ method achieves similar runtime performance to other supervised methods, such as CR-FIQA and eDifFIQA(L). Compared to unsupervised methods, FROQ achieves a far better runtime, even against the least computationally complex GraFIQs, which is around five times slower. Unsurprisingly, LightQnet, which focuses on having a minimal computational footprint, is the fastest method. Overall, in terms of runtime, FROQ resembles supervised techniques, assessing the quality within a single forward pass, consequently achieving excellent performance.  %In conclusion, FROQ resembles supervised methods in the inference runtime, without needing any actual supervised training or to encode additional new information into additional quality-related parameters.

\begin{table}[t]
    \centering
    \caption{\textbf{Results of the Ablation Study.} We perform several ablation studies, thoroughly investigating the effects of individual components on the final experimental result. The table presents the $pAUC$ values, calculated at a discard rate of $20\%$, at FMR$=1e^{-3}$ for the ablation experiments using the AdaFace FR model.
    %The results show pAUC scores at a discard rate of $0.2$, calculated at FMR of $10^{-3}$.
    %The table reports pAUC scores at a discard rate of $0.3$ and a FMR of $10^{-3}$. Average results across all datasets are marked $\overline{\text{pAUC}}$. The best result for each dataset is shown in bold, the overall best result is colored green, the second best blue and the third best red.
    }
    \resizebox{\columnwidth}{!}{%
\begin{NiceTabular}{c  l | l l l l l l l l | r}%[hvlines]

%\Block[]{2-1}{\textbf{{Changes}}} & \Block{2-1}{\textbf{{Adience}}} & \Block{2-1}{\textbf{{LFW}}} & \Block{2-1}{\textbf{{CPLFW}}}  & \Block{2-1}{\textbf{{CFP-FP}}} & \Block{2-1}{\textbf{{CALFW}}}  & \Block{2-1}{\textbf{{AgeDB}}}  & \Block{2-1}{\textbf{{XQLFW}}}  & \Block[2-1]{\textbf{{IJB-C}}} &  \Block{2-1}{\textbf{{IJB-C}}} \\
%\multicolumn{2}{c}{} & {} & {} & {} & {} & {} & {} & {} & {} \\
 %\multicolumn{2}{c}{\textbf{{Changes}}} 
 \Block{2-2}{\textbf{Changes}} & {} 
 & \multirow{2}{*}{{\textbf{Adience}}} & \multirow{2}{*}{{\textbf{LFW}}} & \multirow{2}{*}{{\textbf{CPLFW}}} & \multirow{2}{*}{{\textbf{CFP-FP}}} & \multirow{2}{*}{{\textbf{CALFW}}} & \multirow{2}{*}{{\textbf{AgeDB}}} & \multirow{2}{*}{{\textbf{XQLFW}}} & \multirow{2}{*}{{\textbf{IJB-C}}} & \multirow{2}{*}{{$\overline{pAUC}$}} \\ 
   {} &{} & {} & {} & {} & {} & {} & {} & {} & {} & {} \\
\midrule
\rowcolor{gray!10}
\Block[tikz={preaction={fill, gray!10}}]{1-2} {} & {\textbf{Baseline}} & $0.843$ & $0.754$ & $0.764$ & $0.646$ & $0.925$ & $0.822$ & $0.753$  & $0.798$ & $\mathbf{\underline{0.788}}$ \\
\dashmidrule 
\parbox[t]{2mm}{\multirow{3}{*}{\rotatebox[origin=c]{90}{FIQA}}} & \textbf{eDifFIQA(L)} & $0.892$ & $0.789$ & $0.763$ & $0.634$ & $0.920$ & $0.833$ & $0.777$ & $0.812$ & $0.803$ \\
 & \textbf{CR-FIQA} & $0.891$ & $0.754$ & $0.760$ & $0.651$  & $0.924$ & $0.816$ & $0.771$ &  $0.806$ & $\mathit{\underline{0.797}}$ \\
 & \textbf{CLIB-FIQA} & $0.892$ & $0.789$ & $0.763$ & $0.634$ & $0.920$ & $0.833$ & $0.777$ &  $0.812$ & $0.803$ \\
\dashmidrule 
\parbox[t]{2mm}{\multirow{3}{*}{\rotatebox[origin=c]{90}{w.o. Opt.}}} & \textbf{TOP-$\mathbf{1}$} & $0.848$ & $0.801$ & $0.773$ & $0.593$ &  $0.926$ & $0.844$ & $0.799$ & $0.837$ & $0.803$ \\
 & \textbf{TOP-$\mathbf{5}$} & $0.871$ & $0.769$ & $0.773$ & $0.593$ & $0.926$ & $0.844$ & $0.799$ & $0.809$ & $0.798$ \\
 & \textbf{TOP-$\mathbf{10}$} & $0.873$ & $0.777$ & $0.768$ & $0.593$ & $0.922$ & $0.833$ & $0.785$ & $0.814$ & $\mathit{\underline{0.796}}$ \\
\dashmidrule 
\iffalse
\parbox[t]{2mm}{\multirow{2}{*}{\rotatebox[origin=c]{90}{$S(\cdot)$}}} & \textbf{Std. Dev.} & $0.843$ & $0.754$ & $0.761$ & $0.646$ & $0.751$ & $0.928$ & $0.818$ & $0.797$ & \cellcolor{red!10}$0.787$ \\
 & \textbf{Entropy} & $0.851$ & $0.931$ & $0.888$ & $0.737$ & $0.829$ & $0.966$ & $0.935$ & $1.042$ & $0.897$ \\
\bottomrule
\fi

\multicolumn{2}{l}{\textbf{w.o. BSRGAN}} 
 & $0.851$ & $0.931$ & $0.888$ & $0.737$ & $0.966$ & $0.935$ & $0.763$ & $0.814$ & $\mathit{\underline{0.861}}$ \\

\bottomrule

\end{NiceTabular}
}
        \label{tab:ablation}
            \vspace{-3mm}
\end{table}

\subsection{Ablation Study}\label{sec:experiments_and_results:subsection:ablation_study}

We perform an ablation study to investigate how individual components of the proposed FROQ technique contribute to the final performance, i.e., $(i)$ use of specific auxiliary FIQA techniques, $(ii)$ use of the greedy search algorithm, and $(iii)$ use of the BSRGAN degradation process.

In Table \ref{tab:ablation}, we present the results of the ablation study. The first row marked with \textit{Baseline} presents the results of the FROQ technique; all other rows contain results of the ablation study, separated into the three groups. First, marked with \textit{FIQA}, we present the results of using an alternate auxiliary FIQA technique to produce the pseudo-quality labels. To replace the base auxiliary FIQA, we chose the three best-performing state-of-the-art methods: eDifFIQA(L), CLIB-FIQA, and CR-FIQA. Using CR-FIQA as the auxiliary approach yields the best results, while eDifFIQA(L) and CLIB-FIQA achieve the same averaged result. Surprisingly, all three alternate auxiliary FIQA techniques perform worse than our proposed perturbation-based FIQA approach. Marked with \textit{w.o. Opt.}, we present the results, where we forgo the greedy search for the set of observed intermediate representations $\mathcal{K}$ and instead use the top-$n$ individual representations, specifically the top $1$, $5$, and $10$ layers respectively. We observe that by increasing the number of representations used for the quality assessment task, the results slowly improve, however, they do not reach the performance achieved by the baseline approach. Finally, marked with \textit{w.o.} \textit{BSRGAN}, we present the results obtained using only high and medium quality images from Glint360k, without any additional degradation from BSRGAN. Here, the performance is significantly worse than that of the baseline, alluding to the importance of a wider range of quality values contained in the calibration set. %without applying any degradations to the calibration set of images. of the experiments where a different aggregation function is used, specifically, we focus on the standard deviation (Std. Dev.) and the Entropy of representations. Here, we see that standard deviation slightly outperforms the baseline norm approach, while entropy achieves worse performance.

\begin{figure}[!t]
%\vspace{-5mm}   
    \centering
    \includegraphics[width=0.95\linewidth]{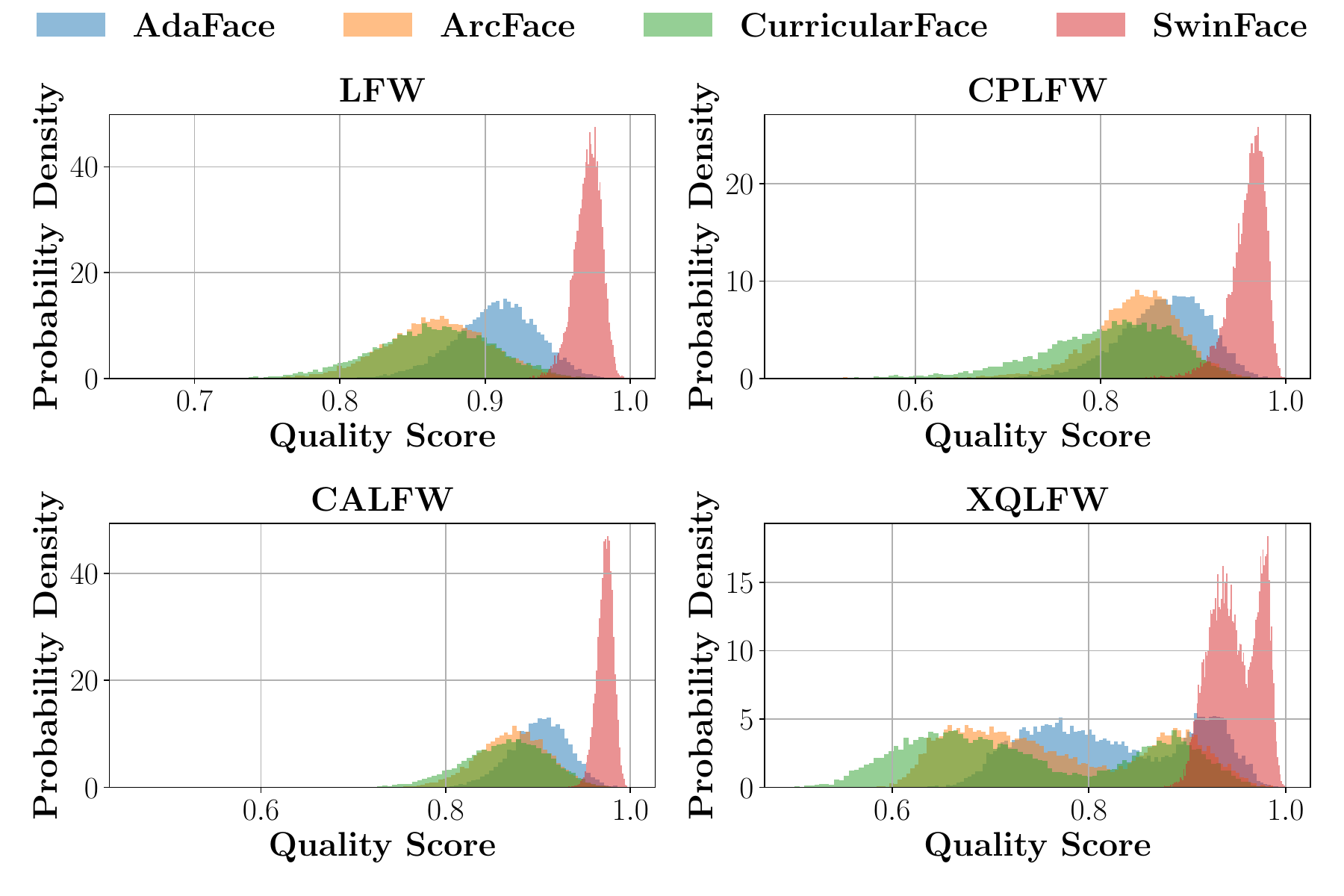}\vspace{-2mm}
    \caption{\textbf{Results of the Qualitative Evaluation.} We evaluate the quality score distributions of the presented technique, using different FR models over four distinct benchmark datasets.}
    \label{fig:qualitative_analysis}\vspace{-2mm}
\end{figure}

\subsection{Qualitative Evaluation}\label{sec:experiments_and_results:subsection:qualitative_evaluation}

In this section, we present the results of the qualitative evaluation of the FROQ technique. In particular, we analyze the distribution of normalized quality scores produced by FROQ when using different FR models. We incorporate all four FR models and four distinct benchmarks, i.e., LFW, CPLFW, CALFW, and XQLFW, into the analysis presented in Fig. \ref{fig:qualitative_analysis}. From the results, differences between FR models can be easily spotted. While AdaFace, ArcFace, and CurricularFace achieve similar distributions, SwinFace exhibits a vastly narrower distribution of quality scores. The disconnect between the models is likely a consequence of the underlying architecture, as the three models are CNN-based, while SwinFace is a Transformer model. %While the distributions of the CNN-based models occupy a wider range, the results are still not optimal, as most quality scores are within a $0.2$ difference.  

\iffalse
\begin{figure}[!h]
    \centering
    \includegraphics[width=0.95\linewidth]{figures/perf.pdf}
    \caption{\textbf{Performance vs. Complexity Comparison over XQLFW}  TODO}
    \label{fig:enter-label}
\end{figure}
\fi

\section{Conclusion}\label{sec:conclusion}

In this paper, we introduced FROQ, a semi-supervised face image quality assessment method that estimates sample quality from intermediate representations within a face recognition (FR) model. Using a greedy search, it selects a subset of informative layers for quality estimation. Extensive experiments on multiple datasets have shown that FROQ outperforms all competing unsupervised FIQA methods and performs similarly to the best supervised techniques without requiring specialized training. %Future work aims to remove the need for labeled data and broaden evaluation to factors like pose and illumination.

%We have presented a new face image quality assessment technique, named FROQ. The main idea behind the technique is to estimate the quality of samples directly from the recognition process, by observing the intermediate representations produced by the given sample. To determine which representations are considered when estimating the quality, the method evaluates all individual layers of a given FR model and constructs the set of intermediate layers $\mathcal{K}$ using a greedy search algorithm. Since the method requires quality labeled images, but no supervised learning, we mark it as a semi-supervised method. The performance of the method far exceeds any previous unsupervised techniques, falling behind only the best supervised methods, which require specialized supervised training and additional parameters to encode quality information. Going forward our plan is to extend the functionality of the technique to work without any labeled data, and expand the scope to other quality-specific factors, such as pose and illumination.
\vspace{2mm} 

\noindent \textbf{Acknowledgments.} Supported by ARIS grants P2-0250, P2-0214, J2-2501 and the Young Researcher Program.

{\small
\bibliographystyle{ieee}
\bibliography{egbib}
}

\end{document}